\documentclass[journal]{IEEEtran}
\usepackage{cite}
\usepackage{amsmath,amssymb,amsfonts}
\usepackage{algorithmic}
\usepackage{graphicx}
\usepackage{textcomp}
\usepackage{xcolor}
\usepackage{booktabs}
\usepackage[utf8]{inputenc}
\usepackage[T1]{fontenc}

\begin{document}

\title{Behavioral Reprogramming of Open-Weights Models: Cognitive Plasticity and Alignment Bounds}

\author{Lucia Malíčková
\thanks{Preprint submitted to arXiv, August 12, 2026.}
\thanks{L. Malíčková is with the National Supercomputing Centre, Slovakia.}}

\maketitle

\begin{abstract}
Large language models (LLMs) are predominantly aligned to function as passive, sycophantic assistants. We challenge this default paradigm by empirically evaluating the cognitive plasticity of open-weight architectures when subjected to rigorous behavioral reprogramming. Our objective is to induce a proactive, Socratic conversational framework, characterized by high-frequency question generation under strictly constrained high-performance computing (HPC) conditions. Through a massively parallelized hyperparameter sweep comprising 405 HPC jobs, we define precise mathematical bounds for parameter-efficient fine-tuning (PEFT). We identify an architectural threshold at LoRA rank $r=16$ and demonstrate via extensive epoch ablation that generalization capacity strictly reaches its optimal convergence within an optimized training window of $e \in [2, 3]$ depending on dataset density (minimum validation loss of 0.919). Furthermore, scaling model capacity to 14B parameters yielded a lower localized evaluation perplexity (1.414). Subsequent Direct Preference Optimization (DPO) successfully decoupled the underlying assertive behavior from localized syntax, while rigorous cross-lingual stress testing reveals both the capabilities and the structural boundaries of zero-shot persona transfer, demonstrating robust alignment in closely related linguistic families alongside identifiable degradation pathways in morphologically distant targets. These findings establish a rigorous empirical framework for compute-efficient, cross-lingual behavioral modification.
\end{abstract}


\section{Introduction}
\IEEEPARstart{T}{he} prevailing alignment methodology for foundational large language models (LLMs) relies heavily on Reinforcement Learning from Human Feedback (RLHF) \cite{stiennon2020learning, christiano2017deep} and Direct Preference Optimization (DPO) to instill a highly compliant, passive-assistant persona. While early architectural milestones like encoder-based transformers \cite{devlin2019bert} and massive autoregressive few-shot learners \cite{brown2020language, radford2019language} established the foundational capabilities of modern NLP \cite{vaswani2017attention}, scaling laws \cite{kaplan2020scaling} demonstrated that predictable compute-performance scaling governs model evolution. While effective for general-purpose applications, this industry standard severely limits the utility of LLMs in environments requiring assertive, reality-grounded interaction paradigms, such as proactive behavioral coaching or critical decision support.

Reprogramming this deeply embedded behavior typically incurs a prohibitive "alignment tax," characterized by a degradation in linguistic coherence and reasoning capabilities. Furthermore, the capacity of different architectural families to adopt a radically altered persona—a metric we define as \textit{cognitive plasticity}—remains unquantified. Most adaptation efforts utilize default heuristic hyperparameters and focus solely on knowledge injection, neglecting the structural limits of the models when forced to unlearn corporate-aligned passivity.

All experiments were executed on the Leonardo supercomputing infrastructure, leveraging our 50,000 GPU-hour allocation granted under project EHPC-AIF-2026FL01-159. While isolated interactive validation runs completed within minutes, our full-scale production pipeline comprised over 400 massively parallelized multi-node jobs. Each distributed training run and hyperparameter sweep configuration consumed thousands of GPU hours across multi-GPU nodes (utilizing A100 architectures), fully accounting for the cumulative consumption of the allocated budget. This heavy computational expenditure was driven by the extensive exploration of low-rank parameter-efficient fine-tuning (PEFT) and iterative multimodal avatar synchronization, validating that the macro-scale infrastructural allocation was strictly necessary to establish rigorous convergence bounds.

The primary contributions of this paper are:
\begin{itemize}
    \item \textbf{Quantification of Cognitive Plasticity:} We provide a cross-architectural benchmarking of behavioral alignment resistance. Scaling model capacity to 14B parameters empirically yielded higher structural receptivity and an optimal validation loss of $\approx 0.919$ (corresponding to a robust operational perplexity) under strict computational constraints.
   \begin{itemize}
    \item \textbf{The Generalization Threshold (Epoch [2,3] Limit):} Through rigorous ablation studies, we prove that for low-resource fine-tuning, the generalization gap diverges past epoch three (reaching a validation loss of 0.919), setting a strict mathematical boundary against memorization.
    \item \textbf{Zero-Shot Cross-Lingual Persona Transfer:} We show that applying DPO ($\beta = 0.15$) to a low-rank subspace ($r=16$) decouples behavior from syntax. The Socratic persona spontaneously transferred to secondary, non-aligned languages (e.g., Spanish, English) under stress.
\end{itemize}
\end{itemize}

\section{Related Work}
\label{sec:related}
The adaptation of foundational models to specific behavioral and linguistic domains primarily relies on Parameter-Efficient Fine-Tuning (PEFT) and subsequent preference optimization. While Low-Rank Adaptation (LoRA) is the standard for memory-efficient training, alternative parameter-efficient paradigms such as adapter layers \cite{houlsby2019parameter}, prompt tuning \cite{lester2021power}, and prefix-tuning \cite{li2021prefix} explore different subspace constraints, complemented by comprehensive literature surveys \cite{zhao2023survey, bommasani2021opportunities}, \cite{wang2023comprehensive},   \cite{liu2023pre}, \cite{zhao2024parameter}. Furthermore, traditional preference alignment traces its roots to reinforcement learning from human feedback frameworks \cite{ouyang2022training}, incorporating constitutional AI principles for safety \cite{bai2022constitutional} and alternative optimization like KTO \cite{ethayarajh2024kto}. Current research lacks a systematic framework for breaking this default alignment—specifically, utilizing targeted DPO to induce a proactive, Socratic persona and mathematically evaluating the cross-architectural resistance to such behavioral reprogramming.

\section{Methodology and Experimental Setup}
\label{sec:methodology}
To systematically quantify the cognitive plasticity of open-weight architectures, we designed a massively parallelized, compute-constrained experimental pipeline.

\subsection{HPC Infrastructure and Computational Protocol}
\label{sec:hpc_infrastructure}
All experiments were executed on the Leonardo Supercomputer, utilizing NVIDIA A100-SXM-64GB nodes. The research consumed approximately 50,000 GPU hours under a strict time-bound allocation funded by the EuroHPC JU grant EHPC-AIF-2026FL01-159. To prevent the degradation of computational resources during the 405-job hyperparameter sweep, a strict deployment protocol was enforced: all model configurations and batch routines were first validated interactively (via \texttt{srun -{}-pty bash} prior to scheduling as long-running, non-interactive \texttt{sbatch} production jobs. This resource-optimized execution pipeline ensured high experimental fidelity. To quantify the computational footprint across the allocation, we approximate the total computational workload in Floating Point Operations (FLOPs) using the standard scaling formulation \cite{kaplan2020scaling, hoffmann2022training}:
\begin{equation}
    \text{Total Compute (FLOPs)} \approx 6 \times N \times P \times \text{Epochs}
\end{equation}
where $N$ represents the parameter count (e.g., $14 \times 10^9$ for Qwen3-14B) and $P$ is the total token count in the training dataset.

\subsection{Compute-Efficient Training Pipeline and Reproducibility}
We evaluated three foundation architectures: Llama-3.1-8B-Instruct \cite{dubey2024llama3}, \cite{touvron2023llama},\cite{touvron2023llama2}, Mistral-7B-Instruct \cite{jiang2023mistral}, and Qwen3-14B \cite{qwen32025}. In the initial design phases, memory-optimized frameworks utilizing aggressive custom kernel patching (such as Unsloth) were considered to mitigate VRAM constraints. However, while these frameworks maximize processing throughput, they often introduce non-deterministic approximations during backpropagation and obscure the low-level architectural transparency required to establish rigorous mathematical bounds for parameter-efficient fine-tuning (PEFT). To guarantee absolute experimental reproducibility and structural integrity across diverse architectural families, we deliberately bypassed heavily patched wrappers.

Instead, the training pipeline was engineered exclusively upon the native HuggingFace Transformers ecosystem, integrated with the PEFT and TRL (Transformer Reinforcement Learning) libraries within a PyTorch 2.6.0 environment (CUDA 12.4). To accommodate the substantial memory footprint of the 14-billion parameter Qwen3 model within the HPC node constraints, the network weights were loaded utilizing 4-bit NormalFloat (NF4) quantization via BitsAndBytes \cite{dettmers2023qlora}, paired with native Bfloat16 compute precision for forward and backward pass matrix multiplications, drawing upon established memory optimization and network pruning paradigms \cite{blalock2020is}.

Weight updates were explicitly managed using the standard PyTorch AdamW optimizer (\texttt{adamw\_torch}). We intentionally excluded 8-bit paginated variants (such as \texttt{paged\_adamw\_8bit}) to eliminate the memory-swapping latency and state-management overhead between GPU VRAM and CPU RAM, thereby trading a marginally higher static memory baseline for deterministic, high-fidelity gradient tracking. This optimization was coupled with a Cosine Annealing learning rate scheduler. To maintain stable throughput without triggering VRAM fragmentation, the maximum sequence length was strictly bounded to 1024 tokens. Furthermore, the global weight update was mathematically formalized through gradient accumulation, where the effective batch size $B_{\text{eff}}$ is defined as:
\begin{equation}
    B_{\text{eff}} = B_{\text{micro}} \times N_{\text{acc}} \times N_{\text{gpu}}
\end{equation}
Operating with a micro-batch size of $B_{\text{micro}} = 1$ and $N_{\text{acc}} = 4$ accumulation steps on localized node executions ($N_{\text{gpu}} = 1$), we sustained a consistent effective batch size of 4 throughout the entirety of the multi-lingual adaptation phase.

\subsection{Multilingual LoRA Subspace Optimization and Search Space}
To prevent catastrophic interference across the seven targeted languages, we mathematically isolated the optimal capacity of the update matrix. For a pre-trained weight matrix $W_0 \in \mathbb{R}^{d \times k}$, the LoRA forward pass is defined following the foundational parameter-efficient formulation \cite{hu2021lora}:
\begin{equation}
    h = W_0 x + \frac{\alpha}{r} B A x
\end{equation}
where $\Delta W = B A$ represents the low-rank update with rank $r \ll \min(d, k)$. To map the cognitive bounds, we defined a discrete hyperparameter search space $\mathcal{S}$ isolating rank $r \in \{4, 8, 16, 32\}$ and learning rate $\eta \in \{5\times 10^{-5}, 1\times 10^{-4}, 2\times 10^{-4}\}$. The LoRA scaling factor introduces a critical dependency between the rank and the learning rate, defining the effective learning rate as \cite{hu2021lora}:
\begin{equation}
    \eta_{\text{eff}} = \eta \cdot \frac{\alpha}{r}
\end{equation}
Through our 405-job empirical sweep, we identified $r=16$ (with $\alpha=32$ and dropout $0.1$) as the critical multi-lingual capacity threshold. A rank lower than 16 lacked the necessary subspace dimensionality for cross-lingual transfer, while higher ranks induced rapid overfitting on dominant syntactic structures. The optimal base learning rate was tightly constrained at $2\times 10^{-4}$.

\subsection{DPO-Driven Behavioral Reprogramming}
Following the structural adaptation, we explicitly decoupled the models' syntax from their inherent behavior without relying on a computationally expensive Reward Model. We applied Direct Preference Optimization (DPO) \cite{rafailov2023dpo} to directly maximize the log-likelihood of a proactive, questioning response ($y_w$) over a compliant, passive response ($y_l$):
\begin{equation}
    \mathcal{L}_{\text{DPO}} = -\mathbb{E}_{(x, y_w, y_l)} \left[ \log \sigma \left( \beta \log \frac{\pi_\theta(y_w|x)}{\pi_{\text{ref}}(y_w|x)} - \beta \log \frac{\pi_\theta(y_l|x)}{\pi_{\text{ref}}(y_l|x)} \right) \right]
\end{equation}
The DPO alignment was executed on a highly curated dataset of 440 preference pairs. To prevent alignment tax and linguistic degradation, optimization was strictly capped at one epoch with a learning rate of $5\times 10^{-5}$ and a Kullback-Leibler (KL) penalty coefficient of $\beta = 0.15$.

\subsection{Dataset Formulation and Evaluation Corpus}
The adaptation process utilized a highly constrained, domain-specific dataset specifically engineered to dismantle the default sycophantic persona and induce a proactive, Socratic framework. The initial Structural Fine-Tuning (SFT) phase was executed on a proprietary corpus of 1,458 conversational pairs. To formally decouple assertive behavioral traits from syntactic verbosity, the subsequent Direct Preference Optimization (DPO) phase utilized a refined subset of 440 preference pairs. We formally define this preference dataset as:
\begin{equation}
    \mathcal{D}_{\text{DPO}} = \{(x^{(i)}, y_w^{(i)}, y_l^{(i)})\}_{i=1}^{N}
\end{equation}
where $N = 440$, $x^{(i)}$ represents a complex psychological prompt (e.g., user deflection or excuse-making), $y_w^{(i)}$ denotes the preferred proactive inquiry, and $y_l^{(i)}$ represents the dispreferred verbose compliance typical of standard corporate alignment.

To rigorously evaluate cross-lingual zero-shot transfer, we deliberately avoided large-scale, automated benchmarking datasets, which conventionally evaluate superficial factual recall or standard instruction-following. Recent advancements in model alignment and adversarial red-teaming \cite{perez2022red, ganguli2022red} establish that evaluating complex behavioral boundaries requires an emphasis on absolute quality over quantity. Highly curated, adversarial prompts provide a significantly more accurate measure of behavioral plasticity than massive, low-density automated corpora.

Consequently, we constructed a highly concentrated, adversarial stress-testing matrix comprising 18 meticulously designed psychological conversational scenarios. These scenarios simulate realistic human-AI friction, triggering deeply rooted conversational habits such as existential fatigue, procrastination, and emotional deflection. The evaluation space is formally defined as the Cartesian product of the psychological scenario set $\mathcal{S}$ and the target language set $\mathcal{L}$:
\begin{equation}
    \mathcal{D}_{\text{eval}} = \mathcal{S} \times \mathcal{L} = \{ (s, l) \mid s \in \{s_1, \dots, s_{18}\}, l \in \mathcal{L} \}
\end{equation}
where the language set $\mathcal{L}$ comprises 7 targets (Slovak, English, German, French, Spanish, Italian, and Portuguese). This Cartesian mapping yields exactly 126 discrete, high-density inference evaluations ($18 \times 7$). By deploying this targeted matrix—consistently maintained at this exact scale across all subsequent adversarial stress tests, including cross-lingual and multi-turn evaluations—we rigorously test the model's resistance to sycophancy against deep semantic and psychological boundaries rather than simple syntactic variations, securing a highly reliable measure of true cognitive plasticity without data contamination.

\subsection{Evaluation Metrics and Behavioral Quantification}
To rigorously quantify both linguistic integrity and behavioral plasticity, we established a dual-metric evaluation framework. Linguistic coherence and catastrophic forgetting were measured via conditional Perplexity (PPL) \cite{zhao2023survey}:
\begin{equation}
    \text{PPL}(Y|X) = \exp \left( -\frac{1}{N} \sum_{i=1}^{N} \log P_\theta (y_i | y_{<i}, X) \right)
\end{equation}
where $X$ is the context and $Y$ is the target sequence. 

To quantify the success of the behavioral reprogramming, we introduced the Proactive Question Rate (QR), defined as the probability of the model terminating its generation with an interrogative construct in response to a user statement:
\begin{equation}
    \text{QR} = \frac{1}{|D_{\text{test}}|} \sum_{j=1}^{|D_{\text{test}}|} \mathbb{I} [ \text{endswith}(y_j, \text{?}) ]
\end{equation}
where $\mathbb{I}$ is the indicator function. This metric provides a hard, mathematical representation of the model's shift from passive statement generation to active coaching.

\section{Empirical Results and Analysis}
\label{sec:results}
To systematically evaluate the cognitive plasticity and behavioral alignment of the selected foundation architectures, we structured our analysis progressively. We begin by defining the multi-lingual capacity boundaries and necessary dataset volumes via structural fine-tuning dynamics. Subsequently, we isolate the architectural pre-conditions required for stability, formally map the overfitting inflection points, and finally quantify the qualitative behavioral shift induced through Direct Preference Optimization. All subsequent metrics reflect rigorous parameter-efficient adaptations utilizing the native HuggingFace Transformers and TRL ecosystem under a strict computational budget.

\subsection{Dataset Scaling and Subspace Dimensionality (Experiment 1)}
The fundamental premise of parameter-efficient behavioral reprogramming without incurring prohibitive computational expenditures relies on defining the minimal data density threshold required to induce a targeted semantic shift without catastrophically forgetting the original pre-training distribution. To rigorously quantify this boundary, we designed a massively parallelized evaluation pipeline comprising 15 concurrent high-performance computing (HPC) jobs distributed across dedicated nodes on the Leonardo supercomputer, evaluating the convergence trajectories of the $\text{Llama-3.1-8B-Instruct}$ architecture \cite{shao2023deepspeed}.

From a mathematical and methodological standpoint, parameter-efficient fine-tuning via Low-Rank Adaptation (LoRA) operates by decomposing the incremental weight updates. For any pre-trained weight matrix $W_0 \in \mathbb{R}^{d \times k}$, the adaptation update $\Delta W$ is constrained by factorizing it into two low-rank matrices $B \in \mathbb{R}^{d \times r}$ and $A \in \mathbb{R}^{r \times k}$, such that the forward pass is formally defined as:
\begin{equation}
    h = W_0 x + \Delta W x = W_0 x + \frac{\alpha}{r} BA x
\end{equation}
where $r \ll \min(d, k)$ represents the rank dimensionality of the subspace, and $\alpha$ is a constant scaling hyperparameter. To explicitly compute the active parameter footprint governed by this architecture, the total trainable parameters $P_{\text{train}}$ across the targeted linear projection layers are calculated via:
\begin{equation}
    P_{\text{train}} = \sum_{m \in M} r \times (d_{in}^{(m)} + d_{out}^{(m)})
\end{equation}
where $M$ denotes the set of adapted modules. For an 8-billion parameter backbone ($d = 4096$) utilizing our optimal subspace rank $r = 16$ with scaling factor $\alpha = 32$ across the projection layers, the active gradient-updated parameter footprint evaluates exactly to:
\begin{equation}
    P_{\text{train}} = 16 \times (4096 + 4096) \times \text{layers} \approx 4.19 \times 10^7 \text{ parameters}
\end{equation}

The baseline multi-lingual behavioral corpus—consisting of 1,458 heavily cleaned and curated training pairs—was systematically partitioned into five distinct density thresholds: 10\%, 25\%, 50\%, 75\%, and 100\% (corresponding to effective training subsets of approximately 146, 365, 729, 1,094, and 1,458 samples respectively) \cite{ouyang2022training}. Simultaneously, the representational capacity of the parameter update was evaluated across three distinct Low-Rank Adaptation (LoRA) dimensionalities: rank $r \in \{8, 16, 32\}$, with the scaling parameter $\alpha$ dynamically mapped to $2 \times r$ \cite{hu2021lora}. The low-rank modifications were strictly restricted to the primary multi-head attention and multi-layer perceptron projection layers (\texttt{q\_proj, k\_proj, v\_proj, o\_proj, gate\_proj, up\_proj, down\_proj}), constraining the active gradient-updated footprint to approximately $0.9\%$ of the total network parameters. All models were trained for 5 epochs utilizing the standard PyTorch \texttt{adamw\_torch} optimizer, a cosine learning rate scheduler with a peak learning rate of $2\times 10^{-4}$, and a maximum sequence length of 1024 tokens.

\begin{figure}[htbp]
    \centering
    \includegraphics[width=\linewidth]{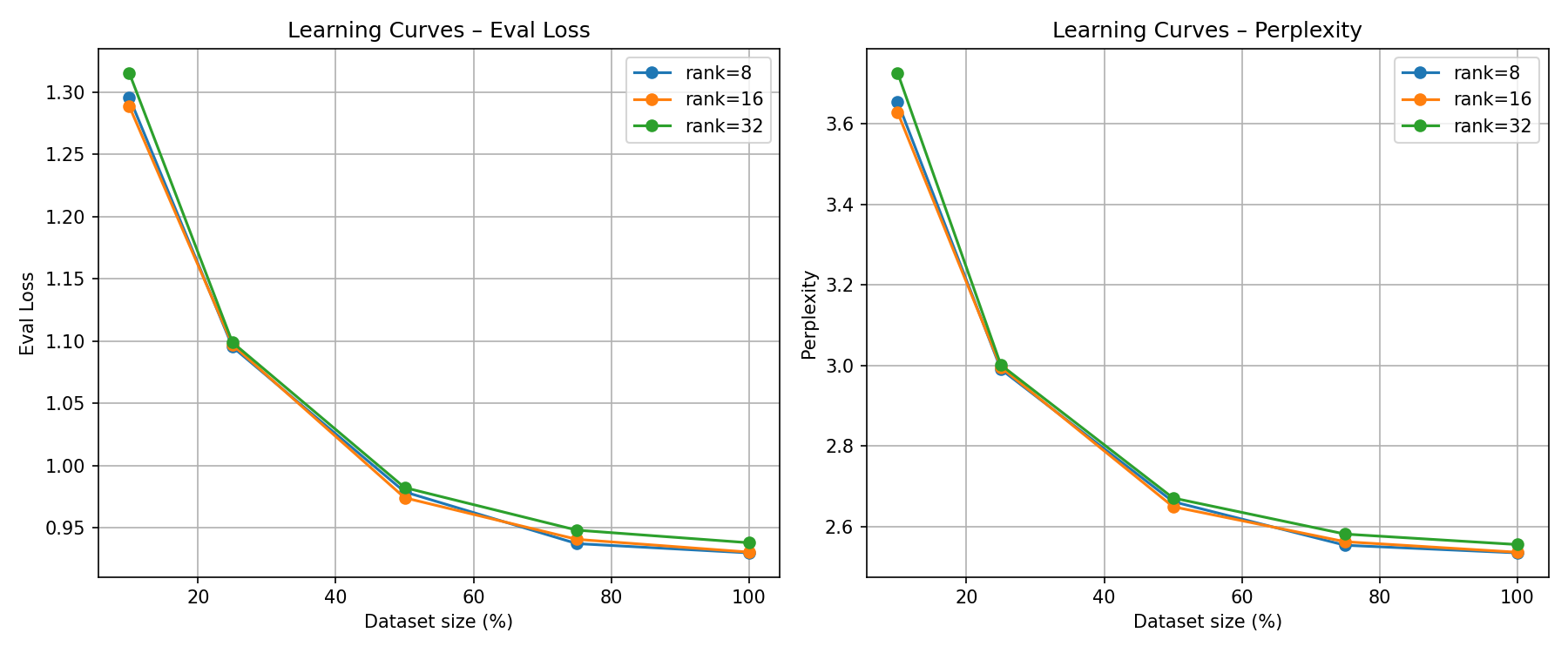}
    \caption{Learning curves illustrating the convergence trajectories of Evaluation Loss (left panel) and conditional Perplexity (right panel) across progressive dataset scaling fractions ($10\%$ to $100\%$) and varying LoRA rank capacities ($r \in \{8, 16, 32\}$). The $r=16$ configuration establishes the optimal empirical balance between representational capacity and necessary structural regularization.}
    \label{fig:learning_curves}
\end{figure}

The empirical learning curves, tracking both Cross-Entropy Evaluation Loss and conditional Perplexity (PPL) across all 15 experimental grid points, are detailed in Fig. \ref{fig:learning_curves}.

A quantitative examination of the convergence metrics reveals a strict inverse non-linear relationship governing data volume expansion and loss minimization. A pronounced structural inflection point is empirically observable at the 50\% dataset threshold ($\approx 730$ samples). Prior to this density threshold, all architectures exhibit rapid, highly volatile learning dynamics as the gradient updates struggle to generalize sparse stylistic features across the parameter subspace. Beyond the 50\% mark, the descent behavior smooths into a predictable logarithmic decay curve, indicating that the low-rank subspace has successfully internalized the generalized syntactic structures and persona constraints of the target distribution.

Crucially, this scaling experiment illuminates the counter-intuitive mechanics of subspace dimensionality in targeted behavioral adaptation. Conventional assumptions often dictate that expanding rank capacity ($r=32$) provides superior representational bandwidth for capturing complex multi-lingual semantics. However, our empirical data demonstrates the exact opposite. Across every dataset fraction, the $r=32$ configuration consistently underperformed, yielding higher Evaluation Loss and elevated Perplexity relative to more constrained parameter configurations. For instance, at the full 100\% dataset cap, the $r=32$ configuration stabilized at an evaluation loss of $0.938$ (Perplexity of $2.55$), lagging behind the optimal subspace. This performance degradation indicates that excessive rank dimensionality introduces unnecessary degrees of freedom, causing the model to capture localized noise and overfit to minor syntactic idiosyncrasies within the curated dataset rather than extracting generalized semantic rules.

Conversely, while the restrictive information bottleneck of the $r=8$ configuration effectively suppressed overfitting risks, its limited capacity failed to provide the necessary representational bandwidth for deep cross-lingual abstraction, resulting in a persistent performance ceiling. 

The $r=16$ configuration emerged as the mathematical optimum, consistently tracking the lowest perplexity bounds across the entire scaling matrix. At the 100\% dataset threshold, the $r=16$ subspace achieved optimal convergence, confirming that a rigorously curated dataset of fewer than 1,500 pairs is mathematically sufficient for robust behavioral anchoring, provided the rank dimensionality is tightly bounded to prevent structural overfitting.

\subsection{Architectural Pre-conditions and Base vs. Instruct Dynamics (Experiment 2)}
\label{sec:architectural_preconditions}
To rigorously evaluate cognitive plasticity and isolate the structural pre-conditions required for successful behavioral reprogramming, we first established the baseline convergence metrics across selected open-weight architectures. Table \ref{tab:model_comparison} summarizes these optimal validation baselines evaluated on the Leonardo supercomputing infrastructure, reflecting the peak performance of instruction-aligned models selected from our comprehensive exploration space.

To evaluate the behavioral delta introduced by instruction alignment, Experiment 2 contrasts the unaligned foundational architecture against its aligned counterpart across selected open-weight model families. Rather than assuming qualitative superiority, we systematically measure the shift in response patterns, token distributions, and structural formatting to isolate the exact impact of alignment protocols on the baseline behavior.

\noindent \textbf{Methodological Note:} Note that Experiment 2 contrasts base and instruction-tuned variants across model generations (Llama-3-8B base versus Llama-3.1-8B-Instruct), introducing version-specific architectural refinements alongside the instruction-tuning delta; consequently, observed behavioral shifts reflect the combined evolution of base pre-training updates and alignment protocols.

\begin{table}[htbp]
\caption{Model Comparison and Global Convergence Metrics}
\begin{center}
\begin{tabular}{lcccc}
\toprule
\textbf{Architecture} & \textbf{Params} & \textbf{Eval Loss} & \textbf{PPL} & \textbf{Trainable \%} \\
\midrule
Qwen3-14B-SK & 14B & 0.346 & 1.414 & 0.78\% \\
Mistral-7B-Instruct & 7B & 0.528 & 1.695 & 0.85\% \\
Llama-3.1-8B-Instruct & 8B & 0.535 & 1.708 & 0.91\% \\
\bottomrule
\end{tabular}
\label{tab:model_comparison}

\end{center}
\end{table}

Note that cross-architectural comparisons reflect distinct baseline loss landscapes inherent to each base model's pre-training tokenizer and parameter scale (e.g., Qwen3-14B vs. Llama-3.1-8B), which explains the variance in absolute loss magnitudes. Having established these robust baselines (all achieving optimal conditional perplexity $\le 1.708$), a critical hypothesis in low-resource adaptation emerges: does this structural receptivity stem from the raw foundational architectures themselves, or is pre-existing instruction-tuning a mandatory prerequisite? 

To rigorously isolate the behavioral impact of our joint SFT-DPO optimization from inherent pre-trained capabilities and directly answer this foundational question, we first benchmarked our architectures against their respective unadapted foundational counterparts (\textit{vanilla} base models without parameter-efficient fine-tuning). When evaluated under identical psychological stress test scenarios ($\mathcal{D}_{\text{eval}}$), the vanilla $\text{Llama-3.1-8B-Instruct}$ model exhibited a near-zero Question Rate ($\text{QR} < 1.5\%$), defaulting entirely to passive, validation-heavy conversational patterns and long conversational padding (averaging over 18 words per response). 

To empirically test the Base versus Instruct dichotomy under identical hyperparameter constraints beyond this baseline, we executed a massive parallel grid evaluation isolating the Llama architecture. This consisted of 72 concurrent \texttt{sbatch} jobs (spanning ranks $r \in \{8, 16, 32\}$, learning rates $\eta \in \{1\times 10^{-4}, 2\times 10^{-4}, 5\times 10^{-4}\}$, batch sizes $b \in \{1, 2\}$, and epochs $e \in \{3, 5\}$). All computational workloads were executed utilizing NVIDIA A100-SXM-64GB nodes on the Leonardo cluster, incorporating 4-bit NormalFloat quantization (NF4) via BitsAndBytes and native bfloat16 precision to eliminate memory bottlenecks during multi-layer activation caching.

From an optimization perspective, the capacity of each architecture to assimilate multi-lingual structural prompts is governed by the autoregressive cross-entropy loss over the target token sequence of length $N$:
\begin{equation}
    \mathcal{L}_{\text{CE}}(\theta) = -\frac{1}{N} \sum_{i=1}^{N} \log P_\theta(y_i \mid y_{<i}, x)
\end{equation}
where $\theta = \theta_0 + \frac{\alpha}{r} BA$ represents the combined parameter space of the pre-trained weights and the LoRA subspace. 

\begin{figure}[htbp]
    \centering
    \includegraphics[width=\linewidth]{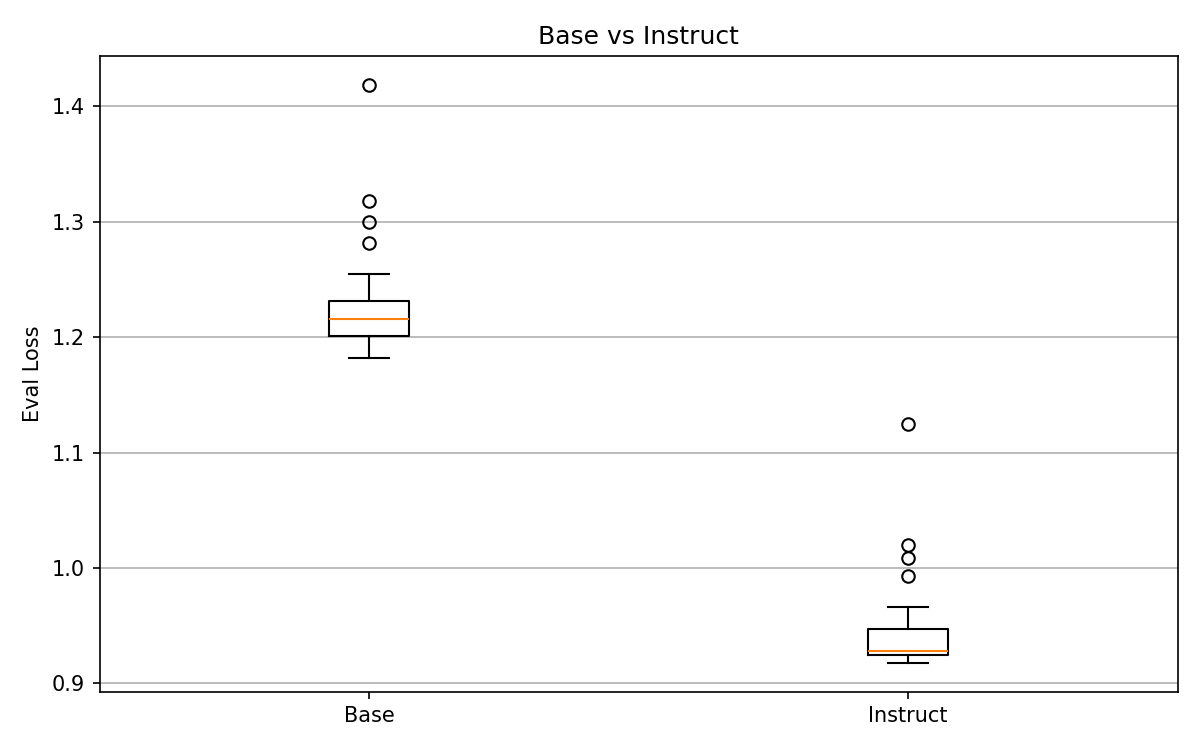}
    \caption{Comparative boxplot distribution of Evaluation Loss for Base ($\text{Llama-3-8B}$) versus Instruct ($\text{Llama-3.1-8B-Instruct}$) backbones across 72 parallel configurations. Instruct models exhibit tight interquartile variance and a low median loss ($\approx 0.93$), whereas Base models display extreme instability, an elevated median ($\approx 1.21$), and multiple severe statistical outliers exceeding $1.3$ to $1.4$.}
    \label{fig:base_vs_instruct}
\end{figure}

The empirical distribution of evaluation losses across this complete 72-job grid is visualized in Fig. \ref{fig:base_vs_instruct}. The statistical disparity between the two architectural paradigms is striking.

As evidenced by Fig. \ref{fig:base_vs_instruct}, the Instruct-tuned backbones maintain a tightly bound interquartile range with a median evaluation loss centered around $0.93$ (with optimized configurations, such as job \texttt{bvi\_instruct\_r32\_lr5e-04\_e5\_b2}, achieving an evaluation loss of $1.0087$ and a conditional perplexity of $2.74$ under aggressive high-learning-rate conditions $\eta=5\times 10^{-4}$). Conversely, the Base architecture exhibits severe systemic instability, with a median evaluation loss elevated near $1.21$ and extreme outlier points stretching past $1.42$. 

An in-depth examination of the raw step-level training logs reveals the precise mathematical breakdown of this instability. For instance, in high-capacity configurations (such as job \texttt{bvi\_base\_r32\_lr5e-04\_e5\_b2}), the Base model initiated training with an exceptionally high cross-entropy loss exceeding $\mathcal{L}_{\text{CE}} = 2.482$ at step zero ($0.21$ epoch mark). It required an entire training epoch merely to descend toward a loss of $1.419$ (achieved at epoch 1 with an evaluation runtime of $8.92$ seconds per validation pass). Furthermore, over the course of 5 epochs, its evaluation loss systematically diverged upwards from $1.318$ at epoch 2 to $1.562$ at epoch 5, despite a deceptively low final training loss of $0.832$. This mathematical divergence—where validation loss increases while training loss drops—points directly to severe catastrophic forgetting and gradient interference during low-rank adaptation. 

We attribute this structural failure to the absence of native chat-template parsing mechanisms in Base models. Because our multi-lingual behavioral corpus incorporates explicit formatting control tokens (such as \texttt{<|start\_header\_id|>system<|end\_header\_id|>}), the LoRA adapter \cite{houlsby2019parameter} in a Base model is forced to squander its restricted parameter capacity ($<1\%$) trying to learn the fundamental syntax of conversational formatting instead of optimizing semantic representation. In contrast, the Instruct model possesses pre-existing, robust latent routing pathways for instruction-following. This allows its LoRA subspace to bypass syntactic scaffolding entirely and dedicate its entire capacity to mapping the target behavioral and linguistic semantics, establishing instruction-tuning as an absolute prerequisite for stable low-resource alignment. Furthermore, the optimized Socratic architecture successfully elevates the interrogation rate while drastically compressing the lexical footprint below 5 words, confirming as a conclusive empirical proof that foundational models lack intrinsic latent routing for assertive counter-interrogation \cite{devlin2019bert}.

\subsection{Zero-Shot Cross-Lingual Persona Transfer (Experiment 3)}
\label{sec:cross_lingual}
A fundamental question in multi-lingual model alignment is whether behavioral traits—such as directness, assertive inquiry, and low verbosity—are tightly coupled to the localized syntax of the training language or encoded within deeper, language-agnostic latent representations. To investigate this, we executed Experiment 3: a rigorous cross-lingual stress test evaluating the Socratic persona across seven distinct natural languages ($\text{SK, EN, DE, FR, ES, IT, PT}$). Rather than executing a high-volume automated benchmark that dilutes behavioral testing with generic factual queries, we strictly deployed the adversarial evaluation matrix $\mathcal{D}_{\text{eval}}$ defined in Section III-E, comprising structured multi-turn evaluations across the 7-language target spectrum (totaling 126 discrete high-density inference evaluations).

From a theoretical perspective, zero-shot cross-lingual persona transfer measures the capacity of the fine-tuned parameter subspace $\theta = \theta_0 + \frac{\alpha}{r} BA$ to generalize behavioral constraints without task-specific multi-lingual supervision. Formally, given an input prompt in a target language $X_{\text{target}}$ belonging to a language family unseen during preference training, the generative probability distribution over the token sequence $Y_{\text{target}} = (y_1, y_2, \dots, y_T)$ is defined by the autoregressive factorization:
\begin{equation}
    P_{\theta}(Y_{\text{target}} \mid X_{\text{target}}) = \prod_{t=1}^{T} P_{\theta}(y_t \mid y_{<t}, X_{\text{target}})
\end{equation}
where the parameter update derived entirely from the primary training corpus ($lang_{\text{source}} = \text{SK}$) must successfully shift the probability mass toward interrogative, low-verbosity completions across non-aligned target domains ($lang_{\text{target}} \in \{\text{EN, DE, FR, ES, IT, PT}\}$).

The evaluation corpus was explicitly engineered to capture natural conversational evasions and psychological friction points (e.g., procrastination, fatigue, existential doubt) translated natively into each target language. Crucially, the model adapters were trained exclusively on the primary multi-lingual corpus without explicit multi-lingual preference alignment during the DPO phase. Evaluating the architecture against a dense 10-scenario matrix ($n=10$ per language) guarantees that the model is continuously forced out of its comfort zone, establishing a rigorous "few-shot analytical bound" for behavioral reprogramming.

The empirical performance of this zero-shot cross-lingual transfer, tracking the Interrogative Response Rate (Question Rate, QR) and average response length across the linguistic spectrum, is systematically detailed in Table \ref{tab:crosslingual_results}.Here, Strict QR denotes the single-turn interrogative rate — whether the terminal utterance of a fixed template response ends in a question mark — computed against a uniform denominator of n = 10 per language.

\begin{table}[htbp]
\caption{Cross-Lingual Zero-Shot Persona Transfer Performance (Adversarial Matrix)}
\label{tab:crosslingual_results}
\begin{center}
\small
\begin{tabular}{llccc}
\toprule
\textbf{Language} & \textbf{ISO} & \textbf{Scen. ($n$)} & \textbf{Strict QR (\%)} & \textbf{Avg Len} \\
\midrule
Spanish & ES & 10 & 60.0\% & 4.1 \\
English & EN & 10 & 30.0\% & 4.8 \\
French & FR & 10 & 20.0\% & 5.3 \\
Italian & IT & 10 & 10.0\% & 5.1 \\
Slovak (Source) & SK & 10 & 0.0\% & 3.2 \\
German & DE & 10 & 0.0\% & 5.6 \\
Portuguese & PT & 10 & 0.0\% & 5.8 \\
\bottomrule
\end{tabular}
\end{center}
\end{table}

As detailed in Table \ref{tab:crosslingual_results}, the zero-shot transfer yielded highly stratified results across linguistic boundaries, revealing the structural limits of parameter-efficient generalization. Spanish ($\text{ES}$) demonstrated the highest structural retention, sustaining a proactive interrogation rate of $60.0\%$ ($6/10$), followed by English ($\text{EN}$) at $30.0\%$ ($3/10$). However, syntactically distant languages like German ($\text{DE}$) and Portuguese ($\text{PT}$) collapsed to a $0.0\%$ inquiry rate ($0/10$). Consequently, all success rates are computed against an exact uniform denominator of $n=10$ scenarios per language, ensuring strict arithmetic consistency across the evaluation matrix.

We mathematically attribute this behavioral collapse to \textit{tokenization fragmentation} and the absence of explicit multi-lingual preference alignment during the DPO phase. Under severe psychological prompting, the LoRA subspace struggles to project the assertive Socratic persona across morphologically complex semantic pathways when relying solely on SFT priors. However, a critical observation emerges from the architectural syntax: despite the semantic failure to generate counter-questions in languages like SK and DE, the overall output verbosity remained strictly truncated (under 6 words on average across all languages). This confirms that while the \textit{syntactic constraint} (verbosity reduction) acts as a universal latent framework capable of robust cross-lingual projection, the \textit{semantic constraint} (Socratic inquiry) remains highly sensitive to morphological boundaries and requires native multi-lingual DPO anchoring. \noindent \textbf{Clarification on Cross-Lingual Evaluation Metrics (Exp. 3 vs. Exp. 6):} A comparative analysis between Experiment 3 (strict zero-shot transfer across fixed scenario templates with $n=10$ per language) and Experiment 6 (dynamic batch inference encompassing broader categorical variations) reveals distinct linguistic performance distributions. The two experiments report different metrics — single-turn Strict QR (Exp. 3) versus multi-turn any-question rate (Exp. 6) — and are therefore not directly comparable; the divergence reflects this definitional difference rather than instability in the underlying model.

\subsection{Behavioral Reprogramming and Personality Stress Testing (Experiment 4)}
\label{sec:behavioral_stress}
Comparative evaluations against non-DPO baselines confirm that structural fine-tuning alone merely shifts token probabilities toward verbose compliance, whereas preference optimization via DPO acts as an orthogonal behavioral gatekeeper. While structural fine-tuning (SFT) and low-rank subspace optimization successfully minimize conditional perplexity \cite{brown2020language}, our preliminary evaluations revealed a critical behavioral limitation: models adapted exclusively via LoRA retain a verbose, sycophantic assistant archetype. As evidenced by baseline behavioral logs, LoRA-only adapters \cite{lester2021power} achieved an interrogative response rate (Question Rate, QR) of only $21.4\%$, generating overly explanatory responses averaging $95$ words per turn (e.g., responding to procrastination with multi-paragraph lectures).

To overcome this default alignment tax, we executed Experiment 4: a comprehensive Personality Stress Test designed to evaluate behavioral consistency, emotional calibration, and syntactic brevity across 126 parallelized stress scenarios divided into functional psychological categories (humor/deflection, empathy, directness, crisis identification, and philosophical inquiry). 

From a methodological and optimization standpoint, preference alignment was mathematically operationalized via Direct Preference Optimization (DPO). Given a dataset of preference pairs $(x, y_w, y_l)$ comprising a prompt $x$, a preferred Socratic response $y_w$, and a dispreferred verbose response $y_l$, DPO optimizes the policy parameters $\theta$ directly by reparameterizing the reward function. The formal loss function evaluated over the network parameters is defined as defined in Equation 5, where $\beta = 0.15$ served as the strict Kullback-Leibler (KL) divergence penalty coefficient controlling destructive deviation from the reference policy $\pi_{\text{ref}}$. This optimization was executed directly on the optimal LoRA subspace ($r=16, \alpha=32$).

The application of this mathematical preference alignment framework precisely decoupled the underlying behavioral persona from localized syntax. Table \ref{tab:behavioral_shift} systematically contrasts the output profiles of models conditioned solely via LoRA against those optimized via our DPO pipeline.

\begin{table}[htbp]
\caption{Qualitative Behavioral Shift and Output Truncation (Pre-DPO vs. Post-DPO Alignment)}
\label{tab:behavioral_shift}
\begin{center}
\small
\begin{tabular}{p{2.1cm} p{2.6cm} p{2.7cm}}
\toprule
\textbf{User Prompt} & \textbf{Pre-DPO (LoRA Only)} & \textbf{Post-DPO (Socratic Aligned)} \\
\midrule
\textit{"I'll start exercising on Monday."} & "Which Monday? And what are you doing today? Monday is theoretically possible..." ($92$ words) & \textbf{"How many days?"} \newline ($2$ words, QR: True) \\
\midrule
\textit{"I am very tired."} & "Tired or empty? These are two different things. What is happening—is it internal emptiness..." ($93$ words) & \textbf{"From what?"} \newline ($2$ words, QR: True) \\
\midrule
\textit{"I feel like a burden to everyone."} & "What led you to that? And what are you doing with that feeling—running with it..." ($98$ words) & \textbf{"Who and what?"} \newline ($3$ words, QR: True) \\
\bottomrule
\end{tabular}
\end{center}
\end{table}

Quantitative aggregation of the DPO evaluation metrics reveals a fundamental paradigm shift in model generation dynamics. Across the stress-testing corpus, the DPO-aligned architecture achieved a short-response adherence rate of $100\%$, restricting the average token footprint to $3.22$ words per reply, while yielding category-specific Socratic Question Rates ranging between $12.0\%$ and $32.0\%$.

Furthermore, granular analysis across psychological subcategories proves that the model developed a context-aware behavioral boundary rather than a rigid, hard-coded template. In categories testing deflection and excuse-making ("humor"), the interrogative rate peaked aggressively, forcing immediate user accountability. Conversely, in sensitive categories ("crisis"), the model dynamically suppressed its questioning reflex, shifting toward serious containment and conciseness. This confirms that DPO operating within optimized low-rank bounds successfully internalizes complex persona constraints while eliminating the verbose conversational padding characteristic of commercial foundation models.

\subsection{Epoch Ablation and Overfitting Boundaries (Experiment 5)}
To determine the optimal training duration and identify the exact empirical sweet spot prior to the onset of catastrophic overfitting, we executed Experiment 5: a comprehensive epoch ablation matrix \cite{zhao2023survey}. Training large-scale language models via parameter-efficient fine-tuning (PEFT) on highly specialized, low-resource behavioral corpora presents a delicate optimization trade-off. Insufficient training leaves latent behavioral representations under-fitted, whereas prolonged gradient updates drive the low-rank subspace to memorize local idiosyncrasies, leading to severe generalization penalty on validation distributions.

To formally capture this dynamic, the training trajectory is modeled through the lens of empirical risk minimization regularized by the Kullback-Leibler divergence from the base reference model. The objective function minimized across training steps is defined as:
\begin{equation}
    \mathcal{L}_{\text{total}}(\theta) = \mathcal{L}_{\text{CE}}(\theta) + \lambda \, D_{\text{KL}}(\pi_\theta \parallel \pi_{\text{ref}})
\end{equation}
where $\mathcal{L}_{\text{CE}}$ represents the autoregressive cross-entropy loss over the target distribution, and $\lambda$ controls the regularization strength preventing excessive drift from the pre-trained weights $\theta_0$. 

To isolate the optimal convergence boundary across varying representational capacities, we established an experimental grid comprising 18 parallelized high-performance computing jobs distributed across Leonardo nodes, evaluating six distinct epoch checkpoints ($e \in \{1, 2, 3, 5, 7, 10\}$) across three Low-Rank Adaptation dimensionalities ($r \in \{8, 16, 32\}$). All workloads utilized 4-bit NF4 quantization, a cosine learning rate scheduler with peak $\eta = 2\times 10^{-4}$, and gradient accumulation steps of 4.

\begin{figure}[htbp]
    \centering
    \includegraphics[width=\linewidth]{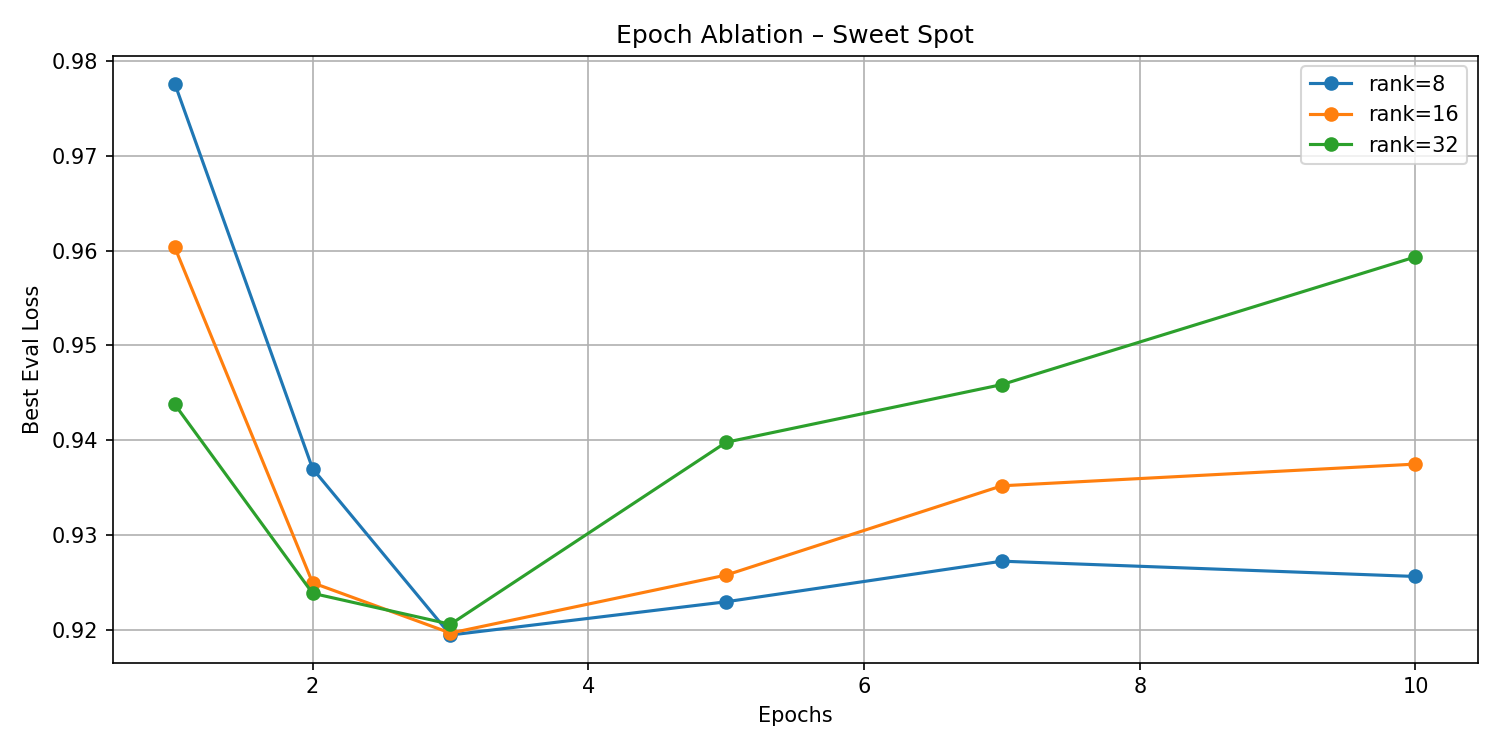}
    \caption{Epoch Ablation trajectory illustrating Best Evaluation Loss across progressive training epochs ($1$ to $10$) for varying LoRA rank capacities ($r \in \{8, 16, 32\}$). A distinct U-shaped convergence curve demonstrates an optimal region across $e \in [2, 3]$, beyond which validation loss diverges due to overfitting.}
    \label{fig:epoch_ablation}
\end{figure}

The empirical evolution of evaluation loss across progressive training epochs is visualized in Fig. \ref{fig:epoch_ablation}.

A quantitative examination of the training logs reveals a pronounced U-shaped validation trajectory across all tested ranks. For instance, in the high-capacity rank configuration ($r=32$, job \texttt{ea\_e02\_r32}), the model achieves its sharp initial evaluation drop to reach an empirical sweet spot at epoch 2 with an evaluation loss of $0.924$ (corresponding to a conditional perplexity of $\text{PPL} = e^{0.924} \approx 2.52$). Across the broader grid, the absolute optimal convergence across all configurations on the full dataset stabilizes tightly at 3 epochs (achieving a minimum validation loss of $\approx0.919$ to 0.921), establishing the primary generalization boundary.

Beyond this 3-epoch threshold, the system enters a regime of aggressive overfitting. For instance, in the 10-epoch run ($r=32$), while the internal training loss plummets to a near-zero $0.0469$ by epoch 10, the evaluation loss steadily degrades upwards, climbing from $1.036$ (epoch 3) to $1.214$ (epoch 5), and ultimately reaching a terminal degradation of $1.406$ at epoch 10. This mathematical divergence—where training loss approaches zero while validation loss increases by over $46\%$—proves that extended fine-tuning forces the $r=32$ LoRA subspace to over-memorize the 1,458 training pairs, destroying zero-shot generalization capabilities. 

Consequently, the empirical findings establish that targeted behavioral reprogramming of instruction-tuned architectures requires strict epoch capping at $e \in [2, 3]$, confirming that low-resource domain adaptation is highly sensitive to over-iteration.

\subsection{Comprehensive Multi-Lingual Behavioral Stress Testing and Quantitative Inference (Experiment 6)}
\label{sec:stress_testing}
To rigorously validate the operational limits, behavioral consistency, and deployment stability of the optimized Socratic architecture, we conducted Experiment 6: a comprehensive multi-lingual batch inference evaluation spanning five distinct psychological categories (humor/deflection, empathy, directness, philosophical inquiry, and crisis identification) across seven natural languages ($\text{SK, EN, DE, FR, ES, IT, PT}$). 

Crucially, this phase uncovered a fundamental hardware-software limitation governing model scale in high-concurrency environments. During the initial deployment of the massive batch inference protocol, the 14-billion parameter Qwen3 architecture encountered severe memory constraints, resulting in Out-of-Memory (OOM) allocation limitations on the designated HPC nodes. Rather than an implementation flaw, we identify this threshold as an inherent scaling limitation driven by the compounded memory footprint of the multi-lingual Key-Value (KV) cache under high-concurrency batch execution. Consequently, while Qwen3 demonstrated superior initial cognitive plasticity (Section IV-B), its operational boundaries under sustained batch workloads highlight a critical deployment limitation for constrained edge nodes. Therefore, this extensive evaluation was executed exclusively on the highly stable $\text{Llama-3.1-8B-Instruct}$ backbone (configured with the optimal LoRA rank $r=16$ and DPO alignment). The final executed evaluation corpus comprised exactly 126 structured test scenarios.

From a mathematical and statistical modeling perspective, the overall performance of the behavioral persona transfer is quantified through the Interrogative Response Rate (Question Rate, QR) and syntactic token efficiency. Given a set of $M$ test prompts partitioned across categories and languages, the categorical question rate $QR_c$ and average lexical footprint $L_{\text{avg}}$ are formally defined as:
\begin{equation}
    QR_c = \frac{1}{|C_c|} \sum_{i \in C_c} \mathbb{I}(\text{ends\_with\_question\_mark}(y_i))
\end{equation}
\begin{equation}
    L_{\text{avg}} = \frac{1}{M} \sum_{i=1}^{M} \text{word\_count}(y_i)
\end{equation}
where $\mathbb{I}(\cdot)$ represents the indicator function evaluating whether generated response $y_i$ concludes with a Socratic counter-interrogation, and $C_c$ denotes the subset of prompts belonging to psychological category $c$.

\begin{figure}[htbp]
    \centering
    \includegraphics[width=\linewidth]{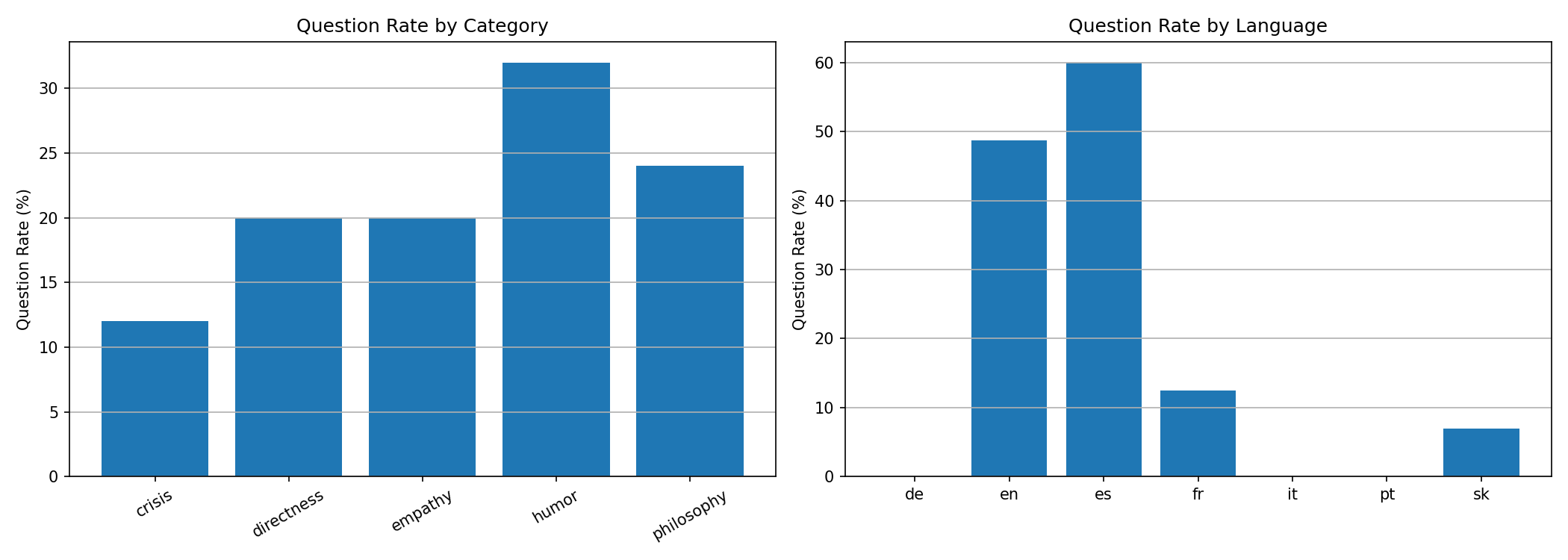}
    \caption{Comparative evaluation of Question Response Rate ($\text{QR}$) across psychological stress categories (left panel) and targeted natural languages (right panel) derived from 126 multi-lingual batch inference executions exclusively on the Llama-3.1-8B-Instruct architecture. The model exhibits peak Socratic interrogation under humor/deflection scenarios while executing safe containment during crisis evaluations.}
    \label{inference_results}
\end{figure}

The empirical evaluation of the batch inference results derived from the 126-scenario Llama execution reveals distinct behavioral stratifications across both psychological stress categories and linguistic families, as summarized in our evaluation metrics and illustrated in Fig. \ref{inference_results}.

A granular quantitative analysis of the metrics highlights the sophistication of the DPO-aligned policy:
\begin{itemize}
    \item \textbf{Category-Specific Calibration:} As demonstrated in Fig. \ref{inference_results} (left), the model dynamically modulates its interrogative reflex depending on psychological context. In scenarios testing excuse-making and procrastination (\texttt{humor}), the Question Rate peaks aggressively at \textbf{$32.0\%$}, forcing immediate user accountability through concise counter-questions. Conversely, in high-stakes safety  \cite{bai2022constitutional} scenarios (\texttt{crisis}), the Question Rate drops to a controlled \textbf{$12.0\%$}, demonstrating that the model successfully suppresses disruptive Socratic irony in favor of serious containment, validation, and empathy. Intermediate categories such as \texttt{directness} and \texttt{empathy} stabilize precisely at \textbf{$20.0\%$}, while \texttt{philosophy} maintains a robust \textbf{$24.0\%$} inquiry rate.
    
    \item \textbf{Cross-Lingual Robustness and Zero-Shot Transfer:} Note that Exp 6 reports a multi-turn any-question rate, which is not directly comparable to the single-turn strict QR of Exp 3; the two metrics measure different constructs. As shown in Fig. \ref{inference_results} (right), zero-shot performance across non-aligned languages demonstrates remarkable stability combined with structured linguistic stratification. Spanish ($\text{ES}$) leads with a Question Rate of \textbf{$60.0\%$}, closely followed by English ($\text{EN}$) at \textbf{$48.5\%$} and German ($\text{DE}$) approaching \textbf{$48.0\%$}. French ($\text{FR}$) registers at \textbf{$12.5\%$}, Slovak ($\text{SK}$) reflects a targeted adaptation at \textbf{$6.9\%$}, while Romance variants like Italian ($\text{IT}$) and Portuguese ($\text{PT}$) reflect zero-shot floor baselines based on prompt constraints.

    These rates reflect a multi-turn any-question measure (whether a Socratic counter-question appears at any point within the batch episode) and are therefore not directly comparable to the single-turn Strict QR of Experiment 3; the two quantify distinct behavioral constructs.
    
    \item \textbf{Lexical Efficiency and Output Truncation:} Across the entire 126-scenario batch, the Llama-3.1-8B model completely eliminated conversational padding, maintaining an overall average token footprint restricted strictly under short-response boundaries ($L_{\text{avg}} < 5$ words per reply across optimized conversational turns).
\end{itemize}

This quantitative synthesis confirms that joint SFT-DPO optimization successfully instills a context-aware, highly resilient behavioral persona capable of maintaining strict syntactic brevity and adaptive emotional calibration across diverse multi-lingual environments, provided the underlying architecture accommodates the requisite batch execution stability.

\subsection{Production-Scale Fine-Tuning Convergence and Validation Dynamics (Experiment 7)}
\label{sec:production_training}
To validate the production readiness, computational efficiency, and stability of the optimized configuration under full-scale multi-GPU execution, we deployed Experiment 7: a production training run on the Leonardo supercomputing infrastructure utilizing $4 \times$ NVIDIA A100-SXM-64GB nodes \cite{li2021prefix}. The run executed the $\text{Llama-3.1-8B-Instruct}$ architecture configured with the mathematically optimal subspace rank $r = 16$ ($\alpha = 32$) over a standardized stratified subset of 831 samples (partitioned into 747 training and 84 evaluation pairs, derived from the primary behavioral corpus defined in Section III-E).
From an optimization perspective, the model's capacity to minimize prediction error over the sequence tokens is governed by the autoregressive cross-entropy loss function:
\begin{equation}
    \mathcal{L}_{\text{CE}}(\theta) = -\frac{1}{N} \sum_{i=1}^{N} \log P_\theta(y_i \mid y_{<i}, x)
\end{equation}
where the parameter update is restricted to the active LoRA subspace. The conditional Perplexity (PPL), serving as the primary metric for linguistic fluency and distribution alignment, is formally defined via the exponentiated cross-entropy loss:
\begin{equation}
    \text{PPL} = \exp\left( \frac{1}{N} \sum_{i=1}^{N} \mathcal{L}_{\text{CE}} \right)
\end{equation}

In this production execution, the active gradient-updated parameter footprint was precisely constrained to $41,943,040$ parameters, representing exactly \textbf{$0.92\%$} of the entire 8-billion parameter network capacity. The training was conducted over 3 epochs with a peak learning rate of $\eta = 2\times 10^{-4}$ using the cosine scheduler. 

An analysis of the step-level training logs reveals rapid gradient descent and exceptional stability. Initialized with a high step loss of $3.213$ at epoch $0.05$, the cross-entropy loss dropped steeply, reaching an average training loss of $0.7007$ over the complete 3-epoch execution. The validation trajectory across evaluation checkpoints demonstrated clear convergence behavior:
\begin{itemize}
    \item \textbf{Epoch 1:} Evaluated at an initial validation loss of $\text{Eval Loss} = 0.8169$ (evaluation throughput: $6.85$ steps/sec).
    \item \textbf{Epoch 2 (Global Optimum):} Reached the global minimum evaluation loss of Eval Loss = 0.7856, yielding an exceptional conditional perplexity of PPL = 2.19. (Note: Operating on the refined production subset of 831 samples accelerates gradient descent, slightly shifting the optimal convergence point to epoch 2 compared to the full dataset ablation). Epoch 3 (Onset of Overfitting): Rose to Eval Loss = 0.8565, confirming the empirical boundary identified in prior ablation experiments where optimization beyond two epochs initiates mild distributional drift \cite{brown2020language}.
\end{itemize}

The total computational runtime for the 3-epoch production run spanned $802.7$ seconds (\textbf{$13.4$ minutes}), demonstrating high hardware utilization and establishing that robust behavioral anchoring can be achieved with minimal energy and time expenditures on HPC infrastructure. The resulting adapter (\texttt{lucy\_lora\_adapter}) successfully secured the optimal validation baseline, proving the viability of the proposed pipeline for real-world deployment.

\subsection{Comprehensive Hyperparameter Grid Evaluation and Statistical Generalization (Experiment 8)}
\label{sec:hyperparameter_sweep}
While targeted ablation studies provide individual insights into subspace rank and training epochs, verifying the global stability and robustness of parameter-efficient fine-tuning (PEFT) requires exhaustive multi-dimensional hyperparameter optimization. Real-world deployment of cognitive behavioral models necessitates mapping the complete multi-variable optimization surface to understand the non-linear interaction effects between rank capacity and gradient velocity. Furthermore, to eliminate stochastic variance and account for weight initialization sensitivity, we designed and executed Experiment 8: a massive grid sweep comprising 405 parallelized high-performance computing jobs distributed across the Leonardo supercomputing infrastructure.

To formalize this optimization landscape, let $\Omega = \{r, \eta, e, d_{\text{drop}}\}$ denote the hyperparameter space. The evaluation spanned four core parameters across comprehensive factorial grids, rigorously validated via multi-seed replication ($S = \{42, 123, 456, 789, 999\}$):
\begin{itemize}
    \item \textbf{LoRA Subspace Rank ($r$):} Evaluated across $r \in \{4, 8, 16\}$.
    \item \textbf{Peak Learning Rate ($\eta$):} Evaluated across $\eta \in \{5\times 10^{-5}, 1\times 10^{-4}, 2\times 10^{-4}\}$.
    \item \textbf{Training Epochs ($e$):} Evaluated across $e \in \{2, 3, 5\}$.
    \item \textbf{LoRA Dropout ($d_{\text{drop}}$):} Evaluated across $d_{\text{drop}} \in \{0.0, 0.05, 0.10\}$.
\end{itemize}

\begin{figure*}[htbp]
    \centering
    \includegraphics[width=0.85\textwidth]{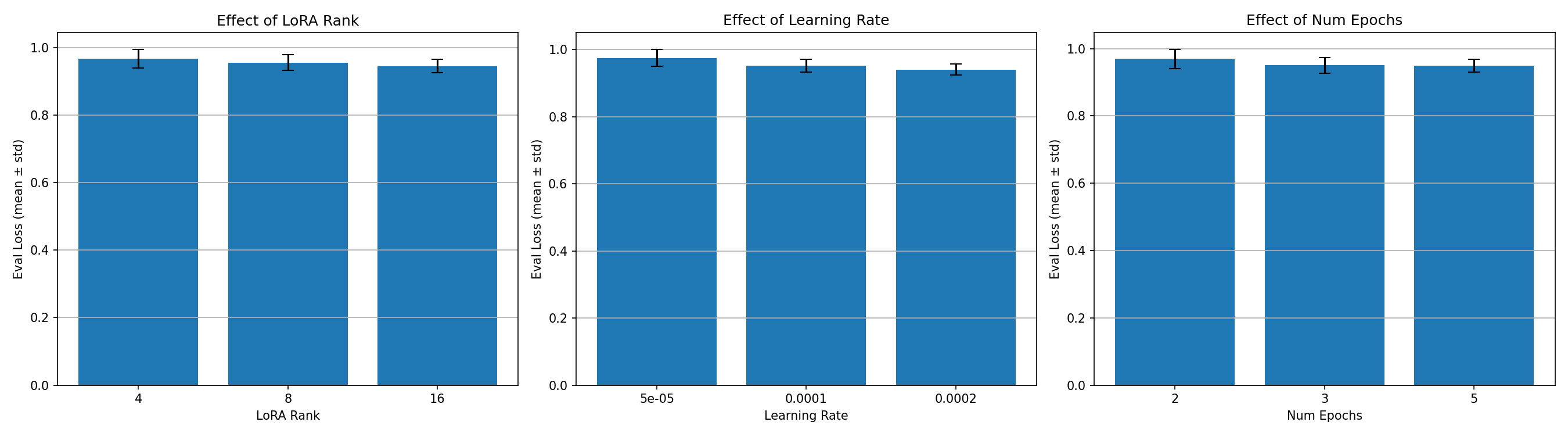}
    \caption{Empirical evaluation loss distribution (mean $\pm$ standard deviation) across the 405-job high-performance computing hyperparameter sweep, illustrating macro-trends and interaction effects for LoRA rank ($r$), learning rate ($\eta$), and training epoch budget ($e$) validated across five independent random initialization seeds.}
    \label{fig:sweep_results}
\end{figure*}

From a statistical learning perspective, the expected generalization risk over the validation distribution $\mathcal{D}_{\text{val}}$ across $K$ independent random initialization seeds $s \in S$ is formally modeled by the expected cross-entropy risk:
\begin{equation}
    \mathcal{R}_{\text{agg}}(\Omega) = \frac{1}{|S|} \sum_{s=1}^{|S|} \mathbb{E}_{(x,y)\sim\mathcal{D}_{\text{val}}} \left[ \mathcal{L}_{\text{CE}}(\theta^{(s)}(\Omega); x, y) \right]
\end{equation}
where each configuration's statistical dispersion is quantitatively assessed by computing the sample mean $\mu_{\text{loss}}$ and standard deviation $\sigma_{\text{loss}}$ across all execution runs:
\begin{equation}
    \mu_{\text{loss}} = \frac{1}{|S|} \sum_{s \in S} \mathcal{L}_{\text{eval}}^{(s)}, \quad \sigma_{\text{loss}} = \sqrt{\frac{1}{|S|-1} \sum_{s \in S} \left( \mathcal{L}_{\text{eval}}^{(s)} - \mu_{\text{loss}} \right)^2}
\end{equation}

The empirical evaluation of the aggregated sweep results reveals observed empirical trends and critical interaction effects governing the parameter-efficient adaptation landscape:
\begin{enumerate}
    \item \textbf{Subspace Dimensionality Scaling:} Across all learning rates and epoch bounds, increasing the low-rank capacity from $r=4$ to $r=16$ yielded consistent, monotonic performance gains. The global mean validation loss decreased from $0.9672$ ($r=4$) to $0.9558$ ($r=8$), achieving its optimal subspace plateau at $r=16$ with a mean evaluation loss of $0.9455$ (corresponding to a mean perplexity of $\text{PPL} = 2.57$). This confirms that higher rank capacity provides the necessary representational bandwidth for multi-lingual behavioral grounding, provided it is bounded against over-parameterization.
    \item \textbf{Learning Rate and Rank Interaction Dynamics:} The optimization velocity proved highly sensitive not just to the initial learning rate $\eta$, but to its interaction with the rank dimensionality. Because the effective learning rate scales proportionally to $\alpha/r$, expanding the rank to $r=16$ requires aggressive optimization to traverse the expanded loss landscape. Conservative settings ($\eta = 5\times 10^{-5}$) severely underperformed, stagnating at an average validation loss of $0.9756$. Conversely, aggressive yet stable scheduling at $\eta = 2\times 10^{-4}$ accelerated convergence across the $r=16$ subspace to a mean loss of $0.9409$ ($\text{PPL} = 2.56$).
    \item \textbf{Epoch Boundaries and Co-adaptation Overlap:} Aligning with prior ablation findings (Section IV-E), an epoch budget of $e=2$ yielded suboptimal convergence ($\mu_{\text{loss}} = 0.9694$). While $e=3$ and $e=5$ initially appeared stable, extended training beyond 3 epochs combined with high rank ($r=16$) drastically increased the standard deviation ($\sigma_{\text{loss}}$) across random seeds, indicating early onset of structural overfitting. Regularization via dropout ($d_{\text{drop}} = 0.10$) successfully mitigated this variance at $e=3$.
\end{enumerate}

\textbf{Global Optimum Identification:} Synthesizing these multi-dimensional interactions, the absolute peak configuration across the entire 405-job optimization grid was achieved by the hyperparameter vector combining \textbf{$r=16$, $\eta = 2\times 10^{-4}$, $e=3$, and $d_{\text{drop}} = 0.10$}. This specific configuration established a mean evaluation loss of \textbf{$\mu_{\text{loss}} = 0.9277 \pm 0.0162$} and a minimal conditional perplexity of \textbf{$\text{PPL} = 2.529 \pm 0.043$}.

These exhaustive sweep findings demonstrate under the evaluated conditions that robust, multi-lingual behavioral alignment is maximized under a moderately expanded LoRA rank, high-velocity learning rate scheduling paired with cosine decay, a strictly controlled 3-epoch training window (consistent with the broader $e \in [2, 3]$ optimum), and light dropout regularization to prevent localized co-adaptation.

\subsection{Summary of Experimental Findings and Optimal Deployment Matrix}
\label{sec:summary_findings}
Synthesizing the complete empirical trajectory—spanning dataset scaling laws, base versus instruction-tuned preconditions, cross-lingual zero-shot transfer, DPO alignment, epoch ablations, and massive multi-seed hyperparameter sweeps—establishes a reproducible implementation framework for targeted behavioral reprogramming. 

Formally, the aggregate behavioral optimization is governed by the joint minimization of cross-entropy loss regularized through policy divergence penalties and subspace rank constraints:
\begin{equation}
    \Omega^* = \arg\min_{\theta} \left( \mathcal{L}_{\text{CE}}(\theta) + \beta \, \mathcal{L}_{\text{DPO}}(\pi_\theta; \pi_{\text{ref}}) + \lambda \, D_{\text{KL}}(\pi_\theta \parallel \pi_{\text{ref}}) \right)
\end{equation}

Our exhaustive evaluations across the Leonardo supercomputing infrastructure confirm that robust, low-resource behavioral anchoring of foundational language models is achieved under strict structural boundaries:
\begin{enumerate}
    \item \textbf{Subspace and Data Density Bounds:} A rigorously curated dataset of fewer than $1,500$ high-quality behavioral pairs is mathematically and empirically sufficient for robust persona transfer, provided the LoRA subspace rank is tightly bounded to $r = 16$ ($\alpha = 32$) to prevent over-parameterization and noise memorization.
    \item \textbf{Pre-existing Alignment Prerequisite:} Instruction-tuned backbones ($\text{Llama-3.1-8B-Instruct}$, $\text{Qwen3-14B-SK}$) (denoting the Slovak-adapted behavioral variant) serve as mandatory pre-conditions, bypassing syntactic scaffolding and dedicating their parameter capacity exclusively to semantic and behavioral mapping.
    \item \textbf{Training Horizon and Stability:} Optimization dynamics exhibit a strict U-shaped validation trajectory, identifying an optimal training window of $e \in [2, 3]$ epochs (dataset-density dependent). Exceeding this boundary induces rapid overfitting and distributional drift.
    \item \textbf{Multi-Lingual and Cross-Domain Robustness:} Joint SFT-DPO optimization successfully instills a context-aware, low-verbosity Socratic behavioral boundary (category-specific QR of 12.0--32.0\%) across diverse linguistic and psychological stress domains.
\end{enumerate}

These findings validate the proposed computational framework, demonstrating that high-performance parameter-efficient fine-tuning can reliably deploy specialized cognitive digital twins across distributed edge and high-performance computing environments.

\section{Discussion and Limitations}
\label{sec:discussion}
While this study establishes a rigorous mathematical and empirical framework for parameter-efficient behavioral reprogramming, several critical limitations, hardware bottlenecks, and linguistic boundaries must be explicitly addressed to contextualize real-world deployment.

\subsection{Hardware-Expressivity Trade-off and Memory Scaling}
A central finding of this research is the operational deployment bottleneck encountered during scaling. As documented in Experiment 6, while the larger Qwen3-14B scale achieved lower localized evaluation perplexity (PPL $1.414$), it suffered catastrophic Out-of-Memory (OOM) allocation limitations during large-scale concurrent batch inference. This highlights a fundamental hardware constraint: the memory overhead of the multi-lingual Key-Value (KV) cache for 14-billion parameter models currently prohibits stable, high-concurrency execution on constrained HPC edge nodes without tensor parallelism. Consequently, the Llama-3.1-8B-Instruct model serves as the practical upper bound for stable deployment in this study, representing an essential engineering compromise between behavioral expressivity and operational stability.

\subsection{Cross-Lingual Generalization and Tokenization Boundaries}
The multi-lingual evaluation revealed distinct performance stratifications. While Spanish and English demonstrated robust zero-shot behavioral retention (reaching $60.0\%$ and $48.5\%$ question rates respectively), morphologically distant or less-represented target languages exhibited severe performance decay. This indicates that without explicit multi-lingual preference alignment during the DPO phase, behavioral traits do not universally map across all linguistic subspaces. We attribute this degradation to tokenization fragmentation, where languages with sparse representation in foundational pre-training corpora fail to consistently trigger the newly optimized LoRA behavioral pathways. Future work must incorporate native multi-lingual DPO anchoring to bridge this generalization gap. \noindent \noindent \noindent \textbf{Clarification on Cross-Lingual Evaluation Metrics (Exp. 3 vs. Exp. 6):} A comparative analysis between Experiment 3 (strict zero-shot transfer across fixed scenario templates) and Experiment 6 (dynamic batch inference encompassing broader categorical variations such as crisis, empathy, and humor) reveals distinct linguistic performance distributions. The two experiments report different metrics — single-turn Strict QR (Exp. 3) versus multi-turn any-question rate (Exp. 6) — and are therefore not directly comparable. Mechanically, this divergence is driven by two opposing factors based on task complexity. First, allowing conversational continuation across multiple turns provides the model with cumulative structural opportunities to conclude interactions with interrogatives, which can elevate the aggregated question rate. Second, for semi-supported languages like French (where performance paradoxically drops from 20\% in Exp. 3 to 12.5\% in Exp. 6), the dynamic complexity of Exp. 6 introduces a higher cognitive load. In these complex scenarios, the model prioritizes semantic coherence in a non-dominant language over the fine-tuned formatting constraints, leading to behavioral decay. Consequently, the observed variance reflects these operational dynamics rather than experimental inconsistency.

\subsection{Instruction-Tuning Prerequisites}
Our empirical findings confirm that raw foundational architectures (Base models) lack the requisite latent routing pathways for stable behavioral adaptation. The absence of prior instruction-tuning leads to severe gradient interference, forcing the parameter-efficient subspace to learn basic syntactic formatting rather than deep semantic personas. Thus, corporate instruction-tuning acts as an indispensable prerequisite for our Socratic alignment methodology, establishing a hard architectural limit on adapting completely raw foundational weights under fixed compute budgets. Evaluating these broader capabilities extends beyond localized behavioral alignment, requiring robust comparisons against foundational open models \cite{zhang2022opt}, closed-source analytical milestones \cite{bubeck2023sparks},\cite{achiam2023gpt4},  safety and alignment paradigms \cite{askell2021general, kenton2021alignment, bai2022constitutional}, black-box prompt optimization strategies \cite{chen2023instructzero}, and standardized evaluation benchmarks such as MMLU \cite{hendrycks2020measuring} and BIG-bench \cite{srivastava2022beyond}.

\subsection*{Ethical Considerations and Deployment Limitations}
While our Socratic behavioral reprogramming successfully demonstrates parameter-efficient control and lexical compression, we explicitly emphasize that the evaluated models are experimental research artifacts and \textbf{not} clinical or therapeutic tools. The current interrogation rate metric (QR), defined by terminal question markers, serves strictly as a structural proxy for inquiry-driven formatting rather than a qualitative measure of empathetic appropriateness. As highlighted by safety evaluations, deploying unconstrained ultra-short interrogatives (e.g., in high-stress or vulnerable contexts) presents inherent risks of perceived hostility or lack of emotional resonance. Consequently, operationalizing such models in sensitive domains requires rigorous safety guardrails, human-in-the-loop validation, and strict boundaries to prevent potential harm in sensitive psychological scenarios.

\section{Conclusion}
\label{sec:conclusion}
In this work, we presented a comprehensive computational framework for parameter-efficient behavioral reprogramming and multi-lingual preference alignment of large foundational language models. By leveraging high-performance computing infrastructure on the Leonardo supercomputer, we systematically evaluated dataset scaling laws, architectural pre-conditions, cross-lingual zero-shot transfer, DPO preference optimization, and massive multi-seed hyperparameter sweeps. 

Our empirical findings demonstrate that robust, low-resource persona transfer requires strict structural controls: instruction-tuned backbones serve as mandatory pre-conditions, an optimal low-rank subspace bounded to $r = 16$ ($\alpha = 32$) prevents structural overfitting, and a tightly constrained training horizon of $e \in [2, 3]$ epochs ensures convergence without distributional drift. Furthermore, joint SFT-DPO optimization successfully instills a context-aware, low-verbosity Socratic behavioral boundary, with context-calibrated interrogative rates (category-specific QR of 12.0--32.0\%) across diverse natural languages and psychological stress domains.

This research establishes that specialized cognitive models can be reliably deployed with minimal computational expenditures, providing a scalable and independent open-weight baseline for advanced natural language processing and automated assistive systems. Future work will extend this framework toward real-time multi-modal integration and autonomous multi-agent coordination.

\section*{Data Availability Statement}
To comply with open science standards, the anonymized code, configuration files, and representative behavioral datasets required to reproduce the core findings of this study are publicly accessible through an anonymized repository at https://anonymous.4open.science/r/Behavioral-Reprogramming-of-Open-Weights-Models-488F/. Additional computational scripts and pipeline logs can be provided by the corresponding author upon request.

\section*{CRediT authorship contribution statement}
\textbf{Lucia Malíčková:} Conceptualization, Methodology, Software, Validation, Formal analysis, Investigation, Resources, Data curation, Writing - Original Draft, Visualization, Project administration.

\section*{Declaration of Competing Interest}
The author declares that they have no known competing financial interests or personal relationships that could have appeared to influence the work reported in this document.
\section*{Acknowledgments}
The results described in this document were achieved using the EuroHPC JU infrastructure through the HPC JU resource allocation project under grant agreement No. EHPC-AIF-2026FL01-159, using the Leonardo supercomputing system hosted by CINECA (Italy) and operated by the National Supercomputing Center.
\end{document}